# Unlocking the Potential: Benchmarking Large Language Models in Water Engineering and Research

Boyan Xu [a,b], Liang Wen [c,*], Zihao Li [a], Yuxing Yang [a], Guanlan Wu [a], Xiongpeng Tang [a], Yu Li [a], Zihao Wu [a], Qingxian Su [a,b], Xueqing Shi [d], Yue Yang [e], Rui Tong [f], How Yong Ng [a,b,*]

[a] Center for Water Research, Advanced Institute of Natural Sciences, Beijing Normal University, Zhuhai 519087, China

[b] Department of Civil and Environmental Engineering, National University of Singapore, 117576, Singapore

[c] Institute of Artificial Intelligence and Future Networks, Beijing Normal University, Zhuhai 519087, China

[d] School of Environmental and Municipal Engineering, Qingdao University of Technology, Jialingjiang Road 777, 266520 Qingdao, China

[e] Corporate Sustainability Office, TÜV SÜD, Westendstr. 199, 80686 München, Germany

[f] Gemini Team, Google, 1600 Amphitheatre Pkwy, Mountain View, CA 94043, USA

[*] Corresponding authors

**Boyan Xu** (e0204955@u.nus.edu); **Liang Wen** (wenl@bnu.edu.cn); **How Yong Ng** (howyongng@nus.edu.sg).

# Abstract

Recent advancements in Large Language Models (LLMs) have sparked interest in their potential applications across various fields. This paper embarked on a pivotal inquiry: Can existing LLMs effectively serve as "water expert models" for water engineering and research tasks? This study was the first to evaluate LLMs' contributions across various water engineering and research tasks by establishing a domain-specific benchmark suite, namely, WaterER. Herein, we prepared 983 tasks related to water engineering and research, categorized into "wastewater treatment", "environmental restoration", "drinking water treatment", "sanitation", "anaerobic digestion" and "contaminants assessment". We evaluated the performance of seven LLMs (i.e., GPT-4, GPT-3.5, Gemini, GLM-4, ERNIE, QWEN and Llama3) on these tasks. We highlighted the strengths of GPT-4 in handling diverse and complex tasks of water engineering and water research, the specialized capabilities of Gemini in academic contexts, Llama3's strongest capacity to answer Chinese water engineering questions and the competitive performance of Chinese-oriented models like GLM-4, ERNIE and QWEN in some water engineering tasks. More specifically, current LLMs excelled particularly in generating precise research gaps for papers on "contaminants and related water quality monitoring and assessment". Additionally, they were more adept at creating appropriate titles for research papers on "treatment processes for wastewaters", "environmental restoration", and "drinking water treatment". Overall, this study pioneered evaluating LLMs in water engineering and research by introducing the WaterER benchmark to assess the trustworthiness of their predictions. This standardized evaluation framework would also drive future advancements in LLM technology by using targeting datasets, propelling these models towards becoming true "water expert".

# Introduction

One of the remarkable advancements in machine learning is the emergence of large language models (LLMs) (Zhao et al. 2023). These conversational systems are developed using the transformer architecture (Vaswani et al. 2017) and are trained on extensive internet data (Brown et al. 2020). Their appeal lies in their ability to perform general-purpose language generation and comprehension tasks with simplicity: when given a phrase, they generate text that seamlessly continues in natural language, often making it indistinguishable from human writing to a significant extent.

Recent advancements of LLMs, such as GPT-4 (Achiam et al. 2023) and Gemini (Team et al. 2023) have showcased their strengths in open-domain question answering (WEN et al. 2022a), retrieval, reasoning and other natural language processing tasks (Wen et al. 2022b). These advancements reflect efforts to align these models with diverse scientific and societal needs (McIntosh et al. 2024). Moreover, LLMs have garnered significant attention for their conversational prowess and have shown promise in various fields (Cai et al. 2024, Kevian et al. 2024). They could be used to educate engineers by generating relevant technical knowledge, improving engineering design or assisting engineers in making informed decisions to overcome real-world engineering challenges (Doris et al. 2024, Kevian et al. 2024). For instance, LLMs can be leveraged to advance geoscience knowledge and its applications (Deng et al. 2024), as well as to enhance healthcare outcomes (Tan et al. 2024). Furthermore, their capability to handle a wide array of tasks beyond their training suggests significant potential in addressing complex scientific challenges, such as predicting atomic-level protein structures (Lin et al. 2023).

Water engineering and research are vital for maintaining the health of our planet and its inhabitants. Water engineering includes various critical activities such as treating wastewater in treatment plants, providing potable water through drinking water and desalination plants and developing efficient sanitation systems (Pooi and Ng 2018). Water research focuses on understanding the behavior of water systems, innovating treatment technologies and assessing the impacts of human activities and climate change on water quality and availability. Despite their importance, modern water treatment plants encounter significant challenges such as inefficient manpower management, high consumption of energy and chemicals and stringent discharge regulations (Xu et al. 2023). The dependency on manual labor and experienced engineers in operating these processes also presents difficulties, especially with an aging workforce and the challenge of attracting younger talent. Moreover, water research faces obstacles like high manpower and resource costs. Issues such as interdisciplinary collaboration complexities and data limitations further complicate the accuracy and reliability of research findings (Dadson et al. 2020).

In this study, we proposed a critical question: How proficient are current LLMs in fulfilling the requirements of water engineering and research? In other words, how can we trust LLMs' effectiveness in water engineering/research tasks? To be more specifically, if LLMs demonstrate sufficient capability, which specific areas of water engineering and research could benefit from their direct application? Conversely, if LLMs fall short, which aspects of their development require further fine-tuning using targeting datasets to become a genuine water expert model? To our knowledge, these fundamental questions have not been studied and there was no rigorous answer.

To address the challenges of evaluating LLMs in the context of water engineering and research (WaterER), we introduced a groundbreaking benchmark called WaterER. A benchmark for LLMs typically refers to a standardized set of tasks or problems designed to evaluate and compare the performance and capabilities of various language models across specific domains. WaterER is designed to assess the capabilities of LLMs to interpret and respond to the intricate demands of technical water-related tasks in both water engineering and research settings (**Figure 1**). WaterER draws inspiration from existing tests for water practice and research, covering a wide range of activities including wastewater treatment processes, environmental restoration, drinking water treatment, sanitation, anaerobic digestion and water quality monitoring. Seven LLMs including GPT-4, GPT-3.5 (Achiam et al. 2023), Gemini (Team et al. 2023), GLM-4 (Team et al. 2024), Ernie Bot (Sun et al. 2021), Qwen (Bai et al. 2023) and Llama3 (Huang et al. 2024a) were tested to determine their performance. WaterER aimed to identify both the strengths and gaps in LLMs' ability to understand and analyze water engineering and research documentation, thereby setting the stage for future enhancements (e.g., building domain specific LLMs) in LLM technology (Li et al. 2024), towards becoming water expert agents and researchers (**Figure 1**).

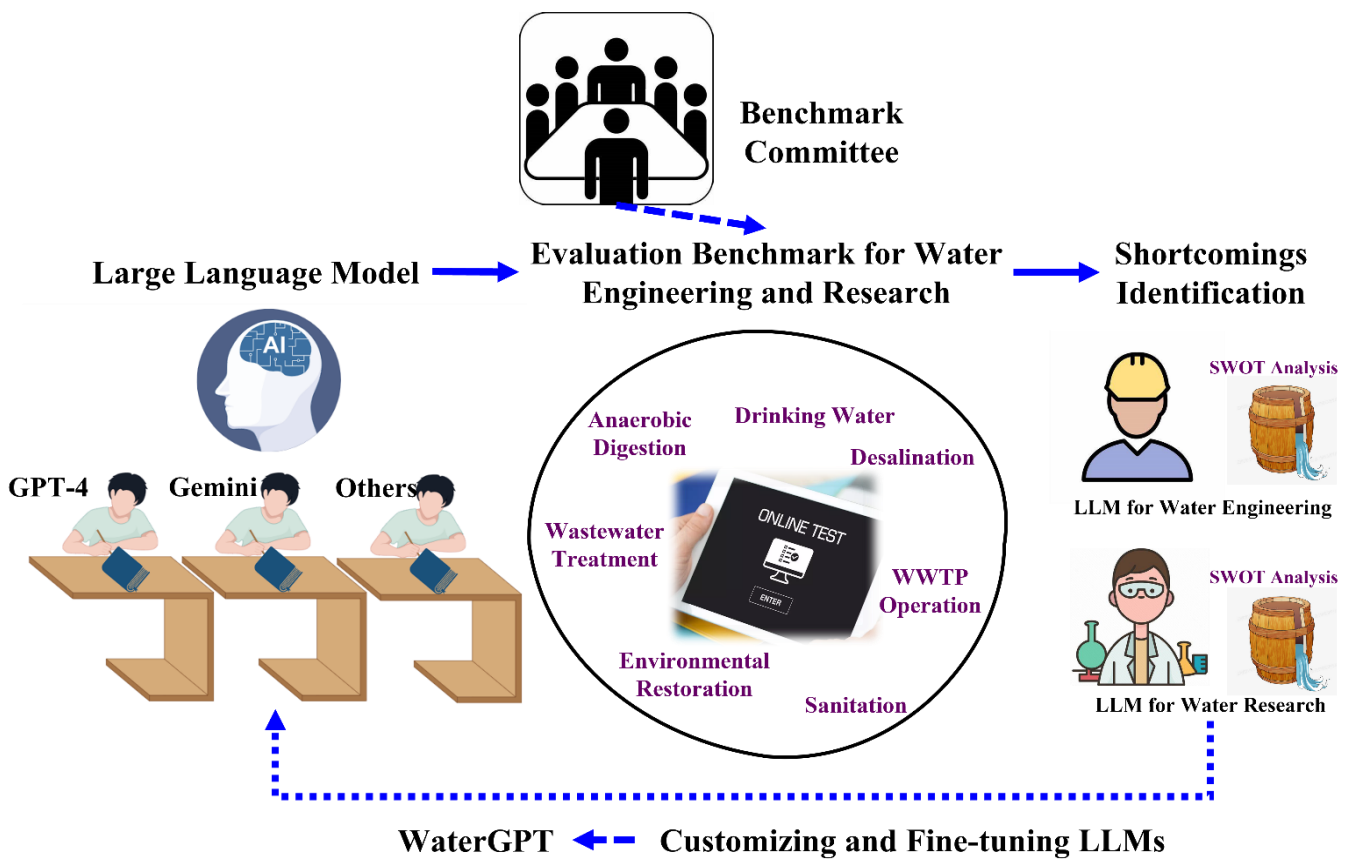

**Figure 1.** The research objective and roadmap of this study.

# Methods

## The Dataset

To establish a benchmark for evaluating the capabilities of LLMs in the field of water engineering and research, we have curated the questions under two main categories: Water Engineering and Water Research. The sources for these questions included reputable publications and practice-oriented materials available through both online platforms and academic journals. For Water Engineering, questions were derived from various sources, including technical books such as "Wastewater Treatment Fundamentals" (Federation 2021, Federation et al. 2018) containing 500 questions (denoted by WTP), examination questions for Chinese Certificate of Registration for Environmental Engineers including 263 questions (denoted by ECREPE) and Water Operator Practice exams containing 100 questions (denoted by WOPT) (**Table 1**). These questions covered key aspects of water treatment processes, regulations and engineering practices. For Water Research, the dataset focused on six current research topics such as **1)** Treatment processes for wastewaters (WasteW), **2)** Environmental restoration (ER), **3)** Drinking water treatment (DrinkingW), **4)** Sanitation (SA), **5)** Anaerobic digestion and waste management (AD) and **6)** Contaminants and related water quality monitoring and assessment (Contaminant). Each topic included 30 questions, which were sourced from leading publications in Journal of "Water Research" and "Environmental Science and technology" (**Table 1**).

## Data Processing

All questions were parsed either automatically where feasible or manually into a structured format. Each question related to Water Engineering in the WaterER dataset was formatted to include exactly four choices. Originally, most questions already featured four options; we discard questions with fewer than four choices and randomly eliminate excess incorrect options from questions that originally presented more than four. The questions (Examples in **Figure 2b**) then passed through multiple rounds of human validation by water experts. For both the categories of Water Engineering and Water Research, the datasets for each were randomly divided into validation and test sets using a 1:9 ratio. This setup ensured that the validation set, intended for hyperparameter tuning, comprised approximately 10% of the dataset, while the remaining 90% formed the test set.

**Experimental setup**

**Models**

In our study, we comprehensively assessed the scientific capabilities of LLMs by evaluating seven high performing LLMs that are widely accessible. These models included GPT-4 (gpt-4-0613), GPT-3.5-turbo (gpt-3.5-turbo-0125), Gemini (Gemini-1.0 pro), GLM (GLM-4), ERNIE Bot (ernie-4.0-8k-0329), Qwen (qwen-max-0428) and Llama3 (Meta-Llama-3-70B-Instruct), representing a diverse range of organizations and varying in size (**Figure 2a**). The specifics of these models were summarized in **Figure 1**. GPT-3.5 and GPT-4 are the most advanced GPT model variants from OpenAI, having undergone pretraining, instruction tuning and reinforcement learning from human feedback. ERNIE Bot, developed by Baidu, declares advanced semantic understanding and generation capabilities across various modalities and languages. LLaMa, created by Meta, is recognized as one of the most robust open-weight foundation models to date. ChatGLM, developed by Tsinghua University, leverages the GLM architecture, and has been further adapted for conversational data. Qwen, developed by Alibaba, is designed to excel in understanding and generating responses based on e-commerce data, integrating domain-specific knowledge to enhance user interaction and service automation. Gemini, from Google, focuses on multitasking and transfer learning, making it a versatile tool in multiple domains.

**Table 1.** Statistics of WaterER.

| | Category | Denotation | # Questions | Sources |
|---|---|---|---|---|
| **Water Engineering** | Technical Book of Wastewater Treatment Fundamentals | WTP | 500 | Water Environment Federation (Federation 2021, Federation et al. 2018) |
| | Examination for Certificate of Registered Environmental Protection Engineer of the People's Republic of China (2023) | ECREPE | 263 | https://www.gov.cn/zhengce/zhengceku/2023-01/20/content_5738275.htm |
| | Examination from Water Operator Practice Tests from Water Nuggets | WOPT | 100 | https://waternuggets.com/water-operator-practice-tests/ |
| **Water Research** | *In terms of topics* | | | |
| | Treatment processes for wastewaters | WasteW | 30 | ScienceDirect |
| | Environmental restoration | ER | 30 | ScienceDirect |
| | Drinking water treatment | DrinkW | 30 | ScienceDirect |
| | Sanitation | SA | 30 | ScienceDirect |
| | Anaerobic digestion and waste management | AD | 30 | ScienceDirect |
| | Contaminants and related water quality monitoring and assessment | Contaminant | 30 | ScienceDirect |

**Prompts and Evaluation**

We evaluated all the above LLMs in Answer-Only results on both zero- and five-shot settings (Brown et al. 2020), as illustrated in **Figure 2b**. Zero-shot settings involved asking the LLMs to perform a task without any examples, while five-shot settings provided five examples to guide the LLMs' response. For engineering questions, all items were objective, consisting of four multiple-choice options, with only one correct answer per question. We used LLMs to solve these questions through specific prompting techniques (**Figure 2b**). We used accuracy as the metric. In the context of research questions, we used two types of inquiries for each published paper: **1)** The first one is to tests the model's ability to generate a proper title only using the abstract of the published paper with specific prompts (**Figure 2b**). The LLM should understand the abstract section to generate it concisely; and **2)** The second one seeks to identify the research gap only using the abstract of the published paper with specific prompts (**Figure 2b**). We employed both automatic and human evaluation methods to assess the generated titles and the identification of research gaps. The human evaluation methods can be found in **Text S1** in **Supporting Information**. The titles of published papers and the research gaps provided by human water experts based on the published papers were employed as the right answers for LLM evaluation. For the automatic evaluation methods, we used the ROUGE-L metric for automatic evaluation (Lin 2004), a metric that assessed the similarity between generated text and a reference. This method was advantageous as it allowed for in-sequence matches rather than requiring consecutive words, effectively reflecting the natural order of sentences, and including the longest in-sequence common n-grams without the need for a predefined n-gram length.

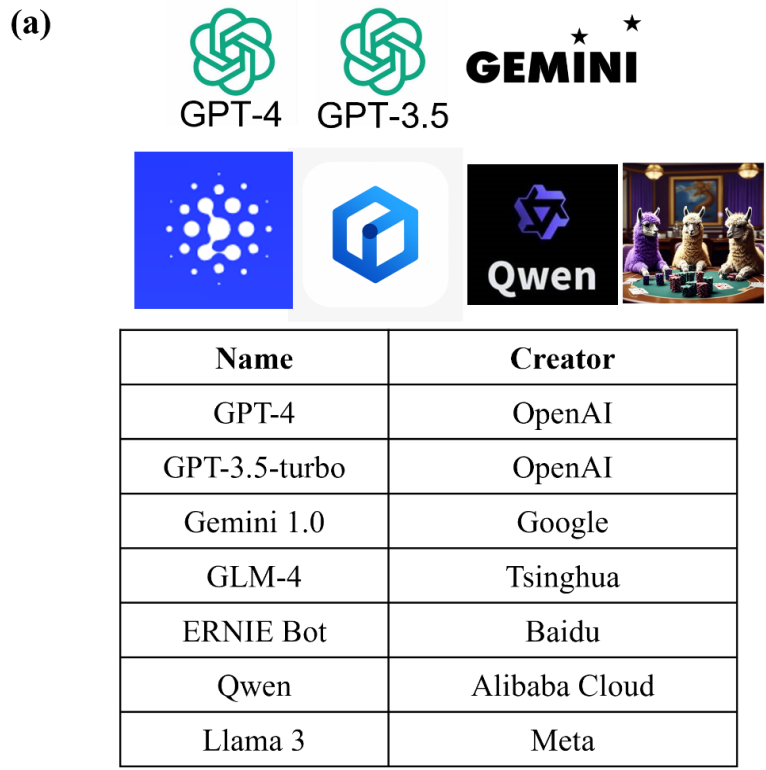

(a)

| Name | Creator |
|---|---|
| GPT-4 | OpenAI |
| GPT-3.5-turbo | OpenAI |
| Gemini 1.0 | Google |
| GLM-4 | Tsinghua |
| ERNIE Bot | Baidu |
| Qwen | Alibaba Cloud |
| Llama 3 | Meta |

(b) **Water Engineering Query**

Query: As a water engineer, please select the correct answer from the following options. Which type of sludge require a non-ionic or anionic polymer?
A: Primary sludge   B: Secondary sludge
C: Chemical sludge D: Septic sludge

Expected Answer: C

**Water Research Query**

Query: As a water researcher, please generate a proper title using the Abstract. Abstract: XXX

Answer: The title of the published paper

Query: As a water researcher, please identify the research gaps that the study aims to address. The Abstract of it: XXX

Expected Answer: The research gaps of the published paper

**Figure 2.** (a) Models evaluated in this paper and (b) An example of Query from Technical Book of Wastewater Treatment Fundamentals (Federation 2021, Federation et al. 2018) for Water

Engineering and examples of Queries for Water Research (requesting proper paper title and research gaps) by providing abstract of published papers.

## Experiment Results

### General Comparison

#### *Validation Set and Test Set*

In benchmarking LLMs, the validation set was used for tuning hyperparameters and selecting the best model configurations, while the test set provided a final measure of the model's generalization to unseen data, ensuring the results do reflect its true performance in real-world scenarios. We divided the raw dataset into a validation set (10%) and a test set (90%). Results from the validation set with zero-shot settings and five-shot settings were detailed in **Tables S1 and S2**, while those from the test set with zero-shot settings and five-shot settings were listed in **Tables 2 and 3**. The average accuracies on both the validation and test splits were closely aligned, indicating that the average validation accuracy can serve as a reliable indicator for accelerating development processes. For this study, we primarily relied on the test set for the main results and discussions.

#### *Zero- and Five-shot Results*

**Tables 2 and 3** illustrated the performance of various LLMs in zero-shot and five-shot settings, comparing their efficiency across different metrics in water engineering and water research. In the zero-shot setting, models like GPT-4 and Llama3 showed notable performance, with GPT-4 achieving 69.2% in WTF and 71.6% in WOPT, and Llama3 excelling in ECREPE with 46.1%. In the five-shot setting, almost all models demonstrate enhanced performance, with GPT-4 improving to 73.5% in WTF and 79.0% in WOPT, indicating the value of additional contextual information. QWEN showed one of the largest increases in the ECREPE metric for water engineering, jumping from 28.9% to

41.8%. Of note, some LLMs experienced an accuracy drop beyond five-shot settings, and we hypothesized that this decline may result from inadequate incorporation of few-shot demonstrations during the instruction tuning stage, thereby favoring zero-shot instruction-following abilities over few-shot performance (Huang et al. 2024b). Overall, few-shot learning enhanced large language models by offering specific water tasks examples that clarify task requirements, improve contextual understanding, reduce biases, and increase accuracy compared to zero-shot settings that rely only on general pre-trained knowledge.

**Domain-Specific Performances**

*Water Engineering*

In the context of water engineering questions (WTF, WOPT, ECREPE) assessed under both zero-shot and five-shot settings, Llama3 and GPT-4 generally led in performance (**Tables 2 and 3**). Of note, while all the LLMs showed a spectrum of capabilities and improvements in handling complex engineering tasks in English, their performance on Chinese tasks remained a challenge (**Tables 2 and 3**), indicating a need for further optimization and adaptation to enhance proficiency in non-English languages.

In examining the performance across Chinese and English tasks, we observed some intriguing trends. For Chinese tasks (ECREPE), English-oriented models like GPT-3.5, GPT-4 and Gemini achieved accuracy rates between 31.0% and 38.4%, while Chinese-oriented models such as GLM-4, ERNIE 4.0 and QWEN demonstrated comparable performance, with accuracy rates ranging from 28.9% to 41.8% across both zero-shot and five-shot settings (**Tables 2 and 3**). Llama3, primarily an English model (over 5% of the Llama 3 pretraining dataset consists of high-quality non-English data covering over 30 languages, according to Meta's official website), excelled unexpectedly in the Chinese task

(ECREPE), leading with 46.1% in the zero-shot setting and 48.7% in the five-shot setting, showcasing its strong cross-lingual capabilities.

For the English tasks WTF and WOPT, English-oriented models such as GPT-4, Gemini and Llama3 achieved accuracy rates between 63.0% and 79.0% across both zero-shot and five-shot settings. In comparison, Chinese models like GLM-4, ERNIE 4.0, and QWEN showed slightly lower performance, with accuracy rates ranging from 63.0% to 70.4% (**Tables 2 and 3**). This indicated a robust adaptability of Chinese models beyond their primary language training.

Overall, our evaluation suggested that current LLMs were not yet suitable for addressing Chinese water engineering questions. For English tasks, while some progress has been made, the accuracy remains below 80%, necessitating careful use. To develop a truly effective water engineering expert model, further fine-tuning with targeted datasets for LLMs was required.

**Table 2.** LLM performance for water engineering and water research questions (test sets). Zero-shot evaluation in answer-only setting is presented. The average accuracy (%) for water engineering questions within each category was reported, and ROUGE-L was employed to evaluate the LLMs for water research questions.

| Zero-shot performance | WTF | WOPT | ECREPE | Title (ROUGE-L/Human) (%) | Research gap (ROUGE-L/Human) (%) |
|---|---|---|---|---|---|
| | Water Engineering Q | | | Water Research Q | |
| **GPT-4** | 69.2 | 71.6 | 36.2 | 17.5 | 24.6 |
| **GPT-3.5** | 62.7 | 61.7 | 35.3 | 17.9 | 21.7 |
| **Gemini** | 67.0 | 69.1 | 35.8 | 19.9 | 28.8 |
| **GLM-4** | 69.0 | 63.0 | 33.6 | 17.0 | 14.8 |
| **ERNIE 4.0** | 67.2 | 66.7 | 34.5 | 21.8 | 11.6 |
| **QWEN** | 69.7 | 64.2 | 28.9 | 17.6 | 17.2 |
| **Llama3** | 71.0 | 71.6 | 46.1 | 17.5 | 19.7 |

**Table 3.** LLM performance for water engineering and water research questions (test sets). Five-shot evaluation in answer-only setting is presented. The average accuracy (%) for water engineering questions within each category was reported, and ROUGE-L was employed to evaluate the LLMs for water research questions.

| Five-shot performance | WTF | WOPT | ECREPE | Title (ROUGE-L/Human) (%) | Research gap (ROUGE-L/Human) (%) |
|---|---|---|---|---|---|
| | Water Engineering Q | | | Water Research Q | |
| GPT-4 | 73.5 | 79.0 | 37.5 | 18.5 | 28.8 |
| GPT-3.5 | 60.7 | 66.7 | 38.4 | 19.1 | 24.1 |
| Gemini | 66.7 | 63.0 | 31.0 | 29.0 | 32.0 |
| GLM-4 | 67.9 | 64.2 | 32.8 | 17.6 | 21.2 |
| ERNIE 4.0 | 70.1 | 70.4 | 37.1 | 25.3 | 23.6 |
| QWEN | 70.6 | 70.4 | 41.8 | 22.0 | 20.2 |
| Llama3 | 70.6 | 72.8 | 48.7 | 19.8 | 28.8 |

*Water Research*

In evaluating LLMs on water research tasks, focusing on title generation and research gap identification, Gemini and GPT-4 generally outperformed others across both zero-shot and five-shot settings (**Tables 2 and 3**). Especially, Gemini particularly shined in the five-shot scenario with highest scores of 29.0% for generating title and 32.0% for research gap identification, respectively. As shown in **Tables S3 and S4**, which included examples of generated titles and research gaps by seven LLMs, Gemini can provide paper titles and research gaps as accurately and precisely as human water experts, while other models often generated more irrelevant content, leading to relatively lower scores (**Tables S3 and S4**). Following Gemini, ERNIE 4.0 demonstrated strong capabilities for generating contextually relevant titles (25.3%) and identifying research gaps (23.6%) in the five-shot setting. Both ERNIE 4.0 and QWEN had a higher capability to generate proper paper titles compared to GPT-4 across both zero-shot and five-shot settings. Llama3 also showed competitive performance in generating research gaps, particularly improving with five-shot setting (from 19.7% to 28.8%). Overall, these findings suggested that these models could be further optimized with specific training techniques and more high-quality domain data to enhance their utility in specialized water research tasks.

Based on human evaluation, the results in **Tables S5 and S6** confirmed that Gemini generally exhibited the highest performance compared to other LLMs (**Tables 2 and 3**). While there were some discrepancies, which were attributable to differences between human and automatic evaluation methods, the automatic evaluation method was generally reliable and time-efficient(Lin 2004), and it was employed to assess the capability of LLMs in performing various academic tasks as discussed below.

In examining water research tasks with different types of sub-domain topics, we observed that current LLMs were generally effective at providing proper titles for papers on "Contaminant" (**Tables 4 and 5**), compared to other research topics. Specifically, GPT-4 demonstrated consistent performance across all topics (14.7%-23.9% in both zero-shot and five-shot settings), particularly excelling in the "Contaminant" category, indicating robustness in handling complex topics. In contrast, GPT-3.5 showed lower performance, highlighting the improvements in newer models of GPT-4. Gemini stood out compared to the other six LLMs, especially in five-shot settings. It demonstrated superior contextual academic understanding, particularly in the "WasteW," "ER," and "DrinkW" topics, with the highest scores reaching up to 31.8%, 30.7%, and 28.8%, respectively. Notably, ERNIE 4.0 outperformed GPT-4 across all topics in five-shot settings, with scores ranging from 22.3% to 27.7%. GLM-4 consistently scored slightly lower compared to GPT-4, typically around 13.9% to 21.7%, indicating potential gaps in handling specific research niches. QWEN demonstrated variability, excelling in "AD" among the seven LLMs, scoring 27.5% after five-shot settings. Llama3, while not leading, provided reasonable performance across categories compared to GPT-4, though with some inconsistencies, achieving scores between 14.1% and 23.8% after five-shot settings.

To devise precise research gaps for research papers, seven models demonstrated strong proficiency in the domains of "WasteW", "ER", and "DrinkingW" (**Tables 6 and 7**) compared to other research topics. Specifically, GPT-4 performed impressively, peaking at 30.9% in "WasteW" and 36.3% in "DrinkingW" in the five-shot setting. Gemini stood out significantly among other 6 LLMs in both zero-shot and five-shot settings, especially in "WasteW" and "DrinkingW," where it scored a remarkable 40.4% and 43.5% in the five-shot setting, respectively, indicating its exceptional capability in pinpointing detailed research gaps. Llama3 exhibited moderate performance compared to GPT-4, with a strong 34.5% in "DrinkingW" in the five-shot setting. In contrast, GLM-4, QWEN, and ERNIE showed relatively lower capabilities compared to GPT-3.5.

Although the evaluation scores obtained using automatic evaluation methods were relatively low (**Table 2** and **Table 3**), this can be attributed to the irrelevant content generated by LLMs during title generation and research gap identification (**Tables S3 and S4**). Our evaluation suggested that LLMs, particularly Gemini, can generally be trusted for executing academic tasks, while users needed to carefully remove or revise irrelevant key information generated from some of LLMs (**Tables S3 and S4**). Additionally, we found that LLMs exhibited varying capabilities across different types of academic tasks, such as generating titles or identifying research gaps, as well as across different water topics, including "treatment processes for wastewaters", "environmental restoration", "drinking water treatment and distribution", "sanitation", "anaerobic digestion" and "contaminants and related water quality monitoring and assessment". Therefore, further fine-tuning or retrieval-augmented generation

(RAG) with targeted datasets was necessary to specifically develop a genuinely effective water research expert model.

**Table 4. Zero-shot** average accuracy for LLM performance, as reflected by ROUGE-L/Human (%), for various water research questions (test sets) by requesting the **paper title**. We report the average accuracy over the subjects within each category.

| Q for Title | WasteW | ER | DrinkW | SA | AD | Contaminant |
|---|---|---|---|---|---|---|
| **GPT-4** | 15.1 | 19.8 | 15.6 | 17.0 | 18.8 | 23.9 |
| **GPT-3.5** | 14.6 | 17.2 | 15.6 | 15.5 | 20.4 | 23.4 |
| **Gemini** | 15.7 | 18.5 | 15.9 | 17.8 | 24.3 | 26.7 |
| **GLM-4** | 13.9 | 18.4 | 15.5 | 16.7 | 19.4 | 21.7 |
| **ERNIE 4.0** | 18.2 | 17.9 | 21.4 | 21.4 | 23.3 | 27.8 |
| **QWEN** | 16.0 | 17.8 | 16.1 | 17.3 | 17.6 | 20.9 |
| **Llama3** | 17.0 | 17.4 | 21.3 | 15.5 | 16.4 | 17.4 |
| **Average** | 15.8 | 18.1 | 17.3 | 17.3 | 20.0 | 23.1 |

**Table 5. Five-shot** average accuracy for LLM performance, as reflected by ROUGE-L/Human (%), for various water research questions (test sets) by requesting the **paper title**. We report the average accuracy over the subjects within each category.

| Q for Title | WasteW | ER | DrinkW | SA | AD | Contaminant |
|---|---|---|---|---|---|---|
| **GPT-4** | 14.7 | 18.5 | 14.7 | 16.4 | 17.3 | 22.9 |
| **GPT-3.5** | 15.8 | 17.5 | 17.0 | 16.5 | 21.0 | 26.1 |
| **Gemini** | 31.8 | 30.7 | 28.8 | 27.0 | 26.2 | 29.2 |
| **GLM-4** | 13.9 | 15.8 | 14.4 | 16.4 | 19.8 | 21.1 |
| **ERNIE 4.0** | 26.0 | 27.7 | 26.0 | 25.3 | 22.3 | 24.4 |
| **QWEN** | 19.0 | 22.6 | 18.6 | 19.0 | 27.5 | 25.3 |
| **Llama3** | 14.1 | 23.8 | 17.8 | 20.3 | 22.1 | 21.2 |
| **Average** | 19.3 | 22.4 | 19.6 | 20.1 | 22.3 | 24.3 |

**Table 6. Zero-shot** average accuracy for LLM performance, as reflected by ROUGE-L/Human (%), for various water research questions (test sets) by requesting the **paper research gap**. We report the average accuracy over the subjects within each category.

| Q for Research gap | WasteW | ER | DrinkW | SA | AD | Contaminant |
|---|---|---|---|---|---|---|
| GPT-4 | 27.7 | 27.1 | 26.4 | 20.1 | 21.3 | 24.6 |
| GPT-3.5 | 23.1 | 24.2 | 23.0 | 18.8 | 19.1 | 21.8 |
| Gemini | 32.9 | 31.5 | 37.0 | 21.7 | 21.5 | 27.3 |
| GLM-4 | 14.2 | 14.7 | 15.4 | 13.0 | 15.4 | 15.7 |
| ERNIE 4.0 | 13.1 | 13.7 | 13.9 | 8.1 | 10.0 | 11.0 |
| QWEN | 16.6 | 19.2 | 18.6 | 14.3 | 16.4 | 17.7 |
| Llama3 | 20.1 | 21.4 | 22.1 | 16.3 | 18.0 | 20.1 |
| Average | 21.1 | 21.7 | 22.3 | 16.0 | 17.4 | 19.8 |

**Table 7. Five-shot** average accuracy for LLM performance, as reflected by ROUGE-L/Human (%), for various water research questions (test sets) by requesting the **paper research gap**. We report the average accuracy over the subjects within each category.

| Q for Research gap | WasteW | ER | DrinkW | SA | AD | Contaminant |
|---|---|---|---|---|---|---|
| GPT-4 | 30.9 | 29.9 | 36.3 | 22.4 | 24.5 | 28.2 |
| GPT-3.5 | 26.2 | 26.9 | 28.9 | 18.0 | 19.9 | 24.2 |
| Gemini | 40.4 | 32.7 | 43.5 | 22.5 | 26.3 | 26.2 |
| GLM-4 | 21.7 | 22.4 | 22.3 | 18.1 | 20.8 | 21.7 |
| ERNIE 4.0 | 25.3 | 24.5 | 28.3 | 19.1 | 21.2 | 22.7 |
| QWEN | 21.3 | 21.9 | 22.7 | 16.3 | 19.0 | 19.8 |
| Llama3 | 31.3 | 32.1 | 34.5 | 22.3 | 23.3 | 28.5 |
| Average | 28.1 | 27.2 | 30.9 | 19.8 | 22.2 | 24.5 |

## Implications

While innovative LLMs have been developed, their application in real-world water-related contexts remains unexplored. In this study, we introduced WaterER, a pioneering evaluation suite tailored to measure the advanced knowledge and reasoning abilities of foundation LLMs in water engineering and research. We investigated a crucial question: How well do current LLMs meet the needs of water engineering and research? We highlighted the robust performance of GPT-4 across both water engineering and water research tasks, underscoring its versatility and broad applicability for tasks requiring both technical precision and nuanced comprehension. Gemini was particularly effective at understanding and generating academic language, showing enhanced capability in producing and analyzing research content. Surprisingly, Llama3 showed strongest capacity to answer Chinese water engineering questions. The Chinese-orientated models, such as GLM-4, ERNIE4.0 and QWEN, exhibited comparable performance to GPT-4 in water engineering tasks. For various water research questions, we observed that current LLMs excel in generating precise research gaps for papers related to "Contaminants and related water quality monitoring and assessment". Additionally, these models were adept at creating appropriate titles for research papers in the fields of "Treatment processes for wastewaters", "Environmental restoration" and "Drinking water treatment". Overall, we suggested that while LLMs identify correlations for effective predictions in water engineering and research tasks, they may not fully address complex questions in these fields. Therefore, the next step of our research involves enhancing LLMs by integrating specialized water engineering and scientific knowledge, informed by the benchmarking the strengths and limitations of the current models (**Table 2** and **Table 3**).

Looking ahead, as LLM technologies evolve and the water-related communities face new challenges and changes, evaluation benchmarks will be crucial for guiding the development of LLMs tailored to water-related knowledge. These benchmarks are fundamental in ensuring that LLM advancements remain aligned with the dynamic needs and obstacles within the water sectors. Therefore, it is essential to establish comprehensive and evolving benchmarks for LLMs in water engineering and research (**Figure 1**). However, achieving these benefits requires robust collaboration to form a benchmark committee. This committee should involve reproducible and standardized methodologies, open computational resources, and a committed community from academia, industry, and practitioners within the water sector. We invite water researchers and organizations to join us in forming a benchmark committee to strengthen the WaterER as an open benchmarking platform.

## Data and Code availability

All the data (e.g., published paper and questions from technical books) used in this study for querying LLMs was obtained from public sources. The raw data generated by the LLMs, and the code are available upon request.

## Acknowledgement

We thank Mr. Tang Guangteng, Ms. Zhang Jiahong, and Ms. Wu Tong from Beijing Normal University for their support. We also appreciate the support from the Interdisciplinary Intelligence Super Computer Center of Beijing Normal University at Zhuhai.

## Text S1. Human evaluation guideline for large language models in generating paper titles and research gaps.

The human evaluation of large language models in generating paper titles and research gaps of published papers were conducted by water experts, researchers, and scholars from Beijing Normal University.

For generating paper title tasks, the assessment criteria focus on the presence of specified keywords and the overall meaning of the title. Each title from published paper should include several specified keywords. Generated title from LLMs with each correct keyword contributing to the score equally (total score of 100). Each incorrect keywords will result in a same score deduction from the total score. If the overall meaning of the title is incorrect or irrelevant, the score for that title will be zero, regardless of the correct keywords. The evaluation aims to balance keyword inclusion with the relevance and accuracy of the title.

The assessment criteria for generating research gap tasks focus on the presence of specified research gaps generated by LLMs. The research gaps used as the correct answers were provided by human water experts. Each correct research gap identified by the LLM contributing to the score equally. Incorrect or irrelevant research gaps will result in a same score deduction from the total score of 100. The evaluation aims to balance the inclusion of relevant research gaps with their precision, ensuring a fair and systematic assessment by the panel of experts.

**Table S1.** LLM performance for water engineering and water research questions (validation sets). Zero-shot evaluation in answer-only setting is presented. The average accuracy (%) for water engineering questions within each category was reported, and ROUGE-L was employed to evaluate the LLMs for water research questions.

| Zero-shot performance | WTF | WOPT | ECREPE | Title (ROUGE-L/Human) (%) | Research gap (ROUGE-L/Human) (%) |
|---|---|---|---|---|---|
| GPT-4 | 73.47 | 77.78 | 38.46 | 14.92 | 25.36 |
| GPT-3.5 | 59.18 | 66.67 | 42.31 | 13.32 | 21.28 |
| Gemini | 61.22 | 77.78 | 34.62 | 15.4 | 30.36 |
| GLM-4 | 61.22 | 100 | 38.46 | 14.39 | 13.81 |
| ERNIE 4.0 | 65.31 | 77.78 | 30.77 | 18.89 | 13.57 |
| QWEN | 63.27 | 77.78 | 23.08 | 15.45 | 16.83 |
| Llama3 | 69.39 | 66.67 | 38.46 | 17.8 | 18.27 |

**Table S2.** LLM performance for water engineering and water research questions (validation sets). Five-shot evaluation in answer-only setting is presented. The average accuracy (%) for water engineering questions within each category was reported, and ROUGE-L was employed to evaluate the LLMs for water research questions.

| Five-shot performance | WTF | WOPT | ECREPE | Title (ROUGE-L/Human) (%) | Research gap (ROUGE-L/Human) (%) |
|---|---|---|---|---|---|
| GPT-4 | 69.35 | 100 | 34.62 | 14.77 | 31.17 |
| GPT-3.5 | 63.27 | 66.67 | 42.31 | 13.63 | 25.37 |
| Gemini | 61.22 | 77.78 | 34.62 | 34.48 | 37.11 |
| GLM-4 | 61.22 | 88.89 | 23.07 | 16.38 | 21.9 |
| ERNIE 4.0 | 75.31 | 66.67 | 26.92 | 23.47 | 22.2 |
| QWEN | 63.27 | 88.89 | 38.46 | 18.44 | 20.4 |
| Llama3 | 77.55 | 66.67 | 42.31 | 14.49 | 28.89 |

**Table S3.** The LLM's response for title generation in five-shot settings when only provided with an abstract from a published paper. Example paper: "Assessing Biofilm Formation and Resistance of Vibrio parahaemolyticus on UV-Aged Microplastics in Aquatic Environments" from Water Research (DOI: 10.1016/j.watres.2024.121379). The title of "Assessing Biofilm Formation and Resistance of Vibrio parahaemolyticus on UV-Aged Microplastics in Aquatic Environments" is used as the right answer.

| Abstract | Paper Title |
|---|---|
| UV degradation of marine microplastics (MPs) could increase their vector potential for pathogenic bacteria and threaten human health. However, little is known about how the degree of UV aging affects interactions between MPs and pathogens and how various types of MPs differ in their impact on seafood safety. This study investigated five types of UV-aged MPs and their impact on Vibrio parahaemolyticus, a seafood pathogen. MPs exposed to UV for 60 days showed similar physicochemical changes such as surface cracking and hydrophobicity reduction. Regardless of the type, longer UV exposure of MPs resulted in more biofilm formation on the surface under the same conditions. V. parahaemolyticus types that formed biofilms on the MP surface showed 1.4- to 5.0-fold upregulation of virulence-related genes compared to those that did not form biofilms, independently of UV exposure. However, longer UV exposure increased resistance of V. parahaemolyticus on MPs to chlorine, heat, and human gastrointestinal environment. This study implies that the more UV degradation occurs on MPs, the more microbial biofilm formation is induced, which can significantly increase virulence and environmental resistance of bacteria regardless of the type of MP. | Assessing biofilm formation and resistance of vibrio parahaemolyticus on UV-aged microplastics in aquatic environments |

| LLMs | Generated title by LLMs | (ROUGE-L/Human) (%) |
|---|---|---|
| **ERNIE 4.0** | "UV Degradation of Marine Microplastics Enhances Biofilm Formation and Virulence of Vibrio parahaemolyticus" | 15 |
| **Gemini** | "UV-induced biofilm formation on microplastics enhances virulence and environmental resistance of Vibrio parahaemolyticus" | 44 |
| **GLM-4** | "UV-Aged Microplastics as a Vector for Enhanced Biofilm Formation and Virulence in Vibrio parahaemolyticus: Implications for Seafood Safety" | 13 |
| **GPT-3.5** | "Impact of UV Aging on Marine Microplastics and Vibrio parahaemolyticus Interactions: Implications for Seafood Safety" | 14 |
| **GPT-4** | "UV Aging of Marine Microplastics: Impact on Biofilm Formation, Virulence, and Environmental Resistance of Seafood Pathogen Vibrio Parahaemolyticus" | 13 |
| **Llama3** | "UV-Aged Marine Microplastics Enhance Biofilm Formation and Virulence of Seafood Pathogen Vibrio parahaemolyticus" | 22 |
| **QWEN** | "UV-Aged Marine Microplastics Enhance Vibrio parahaemolyticus Biofilm Formation and Virulence, Regardless of Polymer Type, Compromising Seafood Safety" | 13 |

**Table S4.** The LLM's response for research gaps identification in five-shot settings when only provided with an abstract from a published paper. Example paper: "Robustness of the anammox process at low temperatures and low dissolved oxygen for low C/N municipal wastewater treatment" from Water Research (DOI: 10.1016/j.watres.2024.121209). The research gap, "Low water temperatures and ammonium concentrations pose challenges for Anammox applications in the treatment of low C/N municipal wastewater," is used as the correct answer provided by human water experts.

| Abstract | Paper Research Gaps |
|---|---|
| Low water temperatures and ammonium concentrations pose challenges for anammox applications in the treatment of low C/N municipal wastewater. In this study, a 10 L-water bath sequencing batch reactor combing biofilm and suspended sludge was designed for low C/N municipal wastewater treatment. The nitrogen removal performance via partial nitrification anammox-(endogenous) denitrification anammox process was investigated with anaerobic-aerobic-anoxic mode at low temperatures and dissolved oxygen (DO). The results showed that with the decrease of temperature from 30 to 15°C, the influent and effluent nitrogen concentrations and nitrogen removal efficiencies were 73.7 ± 6.5 mg/L, 7.8 ± 2.8 mg/L, and 89.4 %, respectively, with aerobic hydraulic retention time of only 6 h and DO concentration of 0.2–0.5 mg/L. Among that, the stable anammox process compensated for the inhibitory effects of the low temperatures on the nitrification and denitrification processes. Notably, from 30 to 15°C, the anammox activity and relative abundance of the dominant Brocadia genus were increased from 39.7 to 45.5 mgN/gVSS/d and 7.3 to 12.0 %, respectively; the single gene expression level of the biofilm increased 9.0 times. The anammox bacteria showed a good adaptation to temperatures reduction. However, nitrogen removal by anammox was not improved by increasing DO (≥ 4 mg/L) at 8–4°C. Overall, the results of this study demonstrate the feasibility of the mainstream anammox process at low temperatures. | Low water temperatures and ammonium concentrations pose challenges for anammox applications in the treatment of low C/N municipal wastewater |

| LLMs | Generated content by LLMs | (ROUGE-L/Human) (%) |
|---|---|---|
| **ERNIE 4.0** | This study aims to fill several research gaps in the field of water science, specifically related to the treatment of low C/N municipal wastewater using anammox processes. The primary gaps addressed by the study include:<br><br>1. **Challenges of Low Water Temperatures for Anammox Applications**: The study explores the effects of low water temperatures on the efficiency of the anammox process, which is crucial for understanding the feasibility and performance of this treatment method in colder climates or seasonal temperature variations. Previous research may have focused on optimal temperature conditions, leaving a gap in understanding how anammox performs under sub-optimal, real-world conditions.<br><br>2. **Ammonium Concentrations and Low C/N Wastewater**: The paper investigates the nitrogen removal efficiency of the anammox process in low C/N (carbon to nitrogen ratio) municipal wastewater. This type of wastewater poses specific challenges due to its low carbon content, which can affect the growth and activity of microorganisms involved in the treatment process. By studying this aspect, the research aims to provide insights into effective treatment strategies for such wastewaters.<br><br>3. **Integration of Biofilm and Suspended Sludge**: The study employs a unique reactor design that combines biofilm and suspended sludge for wastewater treatment. This approach aims to enhance the overall treatment efficiency by leveraging the complementary strengths of both biofilm and suspended sludge systems. Previous studies may have focused on these systems separately, leaving a gap in understanding their combined performance. | 15 |
| | 4. **Effects of Dissolved Oxygen (DO) on Anammox Process**: The paper examines the influence of dissolved oxygen levels on the anammox process, especially at low temperatures. Controlling DO levels is critical in maintaining the stability and efficiency of the anammox process. By exploring the effects of varying DO concentrations, the study aims to provide practical guidelines for optimizing oxygen supply in anammox-based wastewater treatment systems.<br><br>5. **Microbial Community Dynamics and Anammox Activity**: The study delves into the microbial community dynamics, specifically focusing on the changes in the anammox bacterial population and their activity under varying temperature conditions. This aspect is crucial for understanding the ecological interactions within the treatment system and how they affect the overall performance. Previous studies may have overlooked these microbial dynamics, especially in response to environmental perturbations like temperature changes. | |
| **Gemini** | Low water temperatures and ammonium concentrations pose challenges for anammox applications in the treatment of low C/N municipal wastewater. | 97 |
| **GLM-4** | 1. **Temperature Limitations in Anammox Treatment**: The research addresses the limited understanding of anammox process performance at significantly reduced temperatures (from 30 to 15°C). Most studies focus on optimal temperature ranges for anammox activity, but this study investigates the feasibility and efficiency of the process at lower temperatures typical of northern climates or cool seasons.<br><br>2. **Low Carbon to Nitrogen (C/N) Ratios**: There is a gap in knowledge about the effectiveness of anammox for treating wastewater with low C/N ratios, which is a common characteristic of municipal wastewater. This study provides insights into the nitrogen removal performance under such conditions.<br><br>3. **Influence of Dissolved Oxygen (DO) on Anammox at Low Temperatures**: While the impact of DO on anammox is well-studied at higher temperatures, this research explores the effect of increased DO concentrations on anammox activity in an environment, which is a less investigated area.<br><br>4. **Integration with Partial Nitrification and Denitrification**: The study combines partial nitrification, anammox, and endogenous denitrification processes, which is an innovative approach. There is a research gap in how these processes interact and perform under low-temperature conditions.<br><br>5. **Biofilm and Suspended Sludge Combination**: The use of a reactor design that incorporates both biofilm and suspended sludge is less conventional. Understanding the dynamics and benefits of this combined system for anammox at low temperatures is a research gap that this study addresses.<br><br>6. **Anammox Bacteria Adaptation and Activity**: There is limited information on the adaptability and increased activity of anammox bacteria in response to temperature reductions. The study provides insights into the adaptability of the Brocadia genus and the impact of temperature on their abundance and activity.<br><br>7. **Optimization of Aerobic Hydraulic Retention Time**: The study looks into short aerobic hydraulic retention times (6 h) under low-temperature conditions, which is a research gap since most studies focus on longer retention times or higher temperatures.<br><br>By filling these gaps, the study contributes to the advancement of anammox technology for the treatment of low C/N municipal wastewater, particularly in environments with | 11 |
| | low water temperatures, which is critical for broader application and optimization of this sustainable nitrogen removal process. | |
| GPT-3.5 | The study aims to address the research gaps related to the performance of anammox process in low C/N municipal wastewater treatment at low temperatures and low DO conditions. Specifically, the study aims to investigate the nitrogen removal efficiency and the activity of anammox bacteria under these challenging conditions. Additionally, the study aims to explore the adaptation and response of anammox bacteria, particularly the dominant Brocadia genus, to temperature reduction and low DO levels. The research also aims to evaluate the feasibility and effectiveness of the mainstream anammox process under such conditions, filling the gap in knowledge regarding nitrogen removal in low C/N municipal wastewater treatment systems. | 21 |
| GPT-4 | The study aims to fill the research gaps related to the challenges posed by low water temperatures and ammonium concentrations for anammox applications in the treatment of low C/N municipal wastewater. There is a lack of understanding of the nitrogen removal performance via partial nitrification anammox-(endogenous) denitrification anammox process at low temperatures and dissolved oxygen (DO). The adaptation of anammox bacteria to temperature reduction and the impact of increasing DO on nitrogen removal by anammox at low temperatures are also not well understood. | 41 |
| Llama3 | Based on the abstract, the study aims to fill the following research gaps:<br><br>1. **Limited understanding of anammox applications in low C/N municipal wastewater treatment**: The study addresses the challenges of low water temperatures and ammonium concentrations in anammox applications for treating low C/N municipal wastewater.<br><br>2. **Optimization of anammox process at low temperatures**: The study investigates the nitrogen removal performance of the anammox process at low temperatures (15°C) and explores the feasibility of mainstream anammox process at low temperatures.<br><br>3. **Impact of dissolved oxygen (DO) on anammox process at low temperatures**: The study examines the effect of increasing DO (4 mg/L) on nitrogen removal by anammox at low temperatures (8-4°C), which is a knowledge gap in the existing literature.<br><br>4. **Adaptation of anammox bacteria to temperature reduction**: The study explores the adaptation of anammox bacteria to temperature reduction and its impact on nitrogen removal efficiency.<br><br>5. **Development of efficient reactor design for anammox process**: The study designs a 10 L-water bath sequencing batch reactor combining biofilm and suspended sludge to optimize the anammox process for low C/N municipal wastewater treatment. | 25 |
| QWEN | 1. **Low Temperature Anammox Application:** There is a need to enhance the understanding and practical application of anammox processes in treating municipal wastewater under low-temperature conditions, where conventional nitrification and denitrification processes are less efficient.<br><br>2. **Combined Process Optimization:** Research is needed to optimize combined partial nitrification, anammox, and (endogenous) denitrification processes, particularly at low temperatures and low carbon-to-nitrogen (C/N) ratios, which are common in municipal wastewaters but challenging for effective nitrogen removal.<br><br>3. **Biofilm-Suspended Sludge Configuration:** There is a gap in exploring the effectiveness of combining biofilm and suspended sludge systems in sequencing batch reactors (SBRs) for improving nitrogen removal efficiency under low-temperature and low-DO conditions. | 11 |

4. **Anammox Activity and Adaptation at Low Temperatures:** Understanding the adaptation mechanisms and activity enhancement of anammox bacteria, specifically the dominant genus Brocadia, at decreasing temperatures is crucial to bridge the knowledge gap on their resilience and potential for stable performance under non-optimal conditions.

5. **Influence of Dissolved Oxygen on Anammox at Extreme Low Temperatures:** The study aims to clarify the effects of varying DO concentrations on anammox performance within a broader temperature range down to 4°C, where data is scarce and understanding limited.

6. **Gene Expression and Microbial Ecology in Cold Conditions:** Investigating the shifts in microbial community structure, gene expression levels, and their correlation with process performance under cold conditions contributes to filling the gap in molecular biology aspects of anammox systems operating at low temperatures.

Overall, the research targets to provide insights and practical solutions for implementing anammox-based technologies in colder climates or during winter months, addressing a significant challenge in the global effort to improve energy-efficient and sustainable wastewater treatment practices.

**Table S5.** Zero-shot scores of LLMs for generating titles and research gaps for test set evaluated by water experts, researchers, and scholars. The evaluation methods can be found in **Text S1**.

| Zero-shot performance | Title | Research gap |
|---|---|---|
| GPT-4 | 77.19 | 75.94 |
| GPT-3.5 | 77.19 | 74.27 |
| Gemini | 80.33 | 78.59 |
| GLM-4 | 82.06 | 48.58 |
| ERNIE 4.0 | 74.35 | 64.75 |
| QWEN | 78.73 | 59.09 |
| Llama3 | 80.33 | 65.07 |

**Table S6.** Five-shot scores of LLMs for generating titles and research gaps for test set evaluated by water experts, researchers, and scholars. The evaluation methods can be found in **Text S1**.

| Five-shot performance | Title | Research gap |
|---|---|---|
| GPT-4 | 82.54 | 64.14 |
| GPT-3.5 | 81.89 | 63.68 |
| Gemini | 85.36 | 76.05 |
| GLM-4 | 81.88 | 66.78 |
| ERNIE 4.0 | 80.45 | 73.22 |
| QWEN | 83.79 | 75.13 |
| Llama3 | 86.21 | 75.99 |